\documentclass[10pt,twocolumn,letterpaper]{article}

\usepackage{cvpr}              

\usepackage{graphicx}
\usepackage{amsmath}
\usepackage{amssymb}
\usepackage{booktabs}
\usepackage{multirow} 
\usepackage{amssymb}
\usepackage{pifont}
\newcommand{\cmark}{\ding{51}}%
\newcommand{\xmark}{\ding{55}}%
	
\usepackage{soul}

\usepackage[pagebackref,breaklinks,colorlinks]{hyperref}

\usepackage[capitalize]{cleveref}
\crefname{section}{Sec.}{Secs.}
\Crefname{section}{Section}{Sections}
\Crefname{table}{Table}{Tables}
\crefname{table}{Tab.}{Tabs.}

\def\cvprPaperID{4315} 
\def\confName{CVPR}
\def\confYear{2022}

\begin{document}

\title{UBnormal: New Benchmark for Supervised Open-Set Video Anomaly Detection}

\author{Andra Acsintoae$^{1,*}$, Andrei Florescu$^{1,*}$, Mariana-Iuliana Georgescu$^{1,2,3,*}$, Tudor Mare$^{3,}$\thanks{equal contribution\;\;\; $^\diamond$corresponding author}\;,\\
Paul Sumedrea$^{1,*}$, Radu Tudor Ionescu$^{1,3,\diamond}$, Fahad Shahbaz Khan$^{2,4}$, Mubarak Shah$^{5}$\\
$^1$University of Bucharest, Romania, $^2$MBZ University of Artificial Intelligence, UAE\\
$^3$SecurifAI, Romania, $^4$Link\"{o}ping University, Sweden, $^5$University of Central Florida, US\vspace*{-0.3cm}
}
\maketitle

\begin{abstract}
\vspace*{-0.2cm}
Detecting abnormal events in video is commonly framed as a one-class classification task, where training videos contain only normal events, while test videos encompass both normal and abnormal events. In this scenario, anomaly detection is an open-set problem. However, some studies assimilate anomaly detection to action recognition. This is a closed-set scenario that fails to test the capability of systems at detecting new anomaly types. To this end, we propose UBnormal, a new supervised open-set benchmark composed of multiple virtual scenes for video anomaly detection. Unlike existing data sets, we introduce abnormal events annotated at the pixel level at training time, for the first time enabling the use of fully-supervised learning methods for abnormal event detection. To preserve the typical open-set formulation, we make sure to include disjoint sets of anomaly types in our training and test collections of videos. To our knowledge, UBnormal is the first video anomaly detection benchmark to allow a fair head-to-head comparison between one-class open-set models and supervised closed-set models, as shown in our experiments. Moreover, we provide empirical evidence showing that UBnormal can enhance the performance of a state-of-the-art anomaly detection framework on two prominent data sets, Avenue and ShanghaiTech. Our benchmark is freely available at \small{\url{https://github.com/lilygeorgescu/UBnormal}}.
\vspace*{-0.3cm}
\end{abstract}

\setlength{\abovedisplayskip}{3.5pt}
\setlength{\belowdisplayskip}{3.5pt}

\section{Introduction}
\label{sec:intro}

In spite of the growing interest in video anomaly detection \cite{Dong-Access-2020,Doshi-CVPRW-2020a,Georgescu-CVPR-2021,Georgescu-TPAMI-2021,Gong-ICCV-2019,Ionescu-WACV-2019,Ionescu-ICCV-2017,Ji-IJCNN-2020,Lee-TIP-2019,Liu-ICCV-2021,Lu-ECCV-2020,Nguyen-ICCV-2019,Pang-CVPR-2020,Park-CVPR-2020,Ramachandra-WACV-2020a,Ramachandra-PAMI-2020,Smeureanu-ICIAP-2017,Sun-ACMMM-2020,Wang-ACMMM-2020,Wu-TNNLS-2019,Yu-ACMMM-2020,Zaheer-CVPR-2020}, which generated significant advances leading to impressive performance levels \cite{Georgescu-CVPR-2021,Georgescu-TPAMI-2021,Ionescu-CVPR-2019,Lee-TIP-2019,Liu-ICCV-2021,Tang-PRL-2020,Vu-AAAI-2019,Wang-ACMMM-2020,Yu-ACMMM-2020,Zaheer-CVPR-2020,Zaheer-ECCV-2020}, the task remains very challenging. The difficulty of the task stems from two interdependent aspects: $(i)$ the reliance on context of anomalies, and $(ii)$ the lack of abnormal training data. The former issue can be explained through a simple comparative example considering a truck driven on a street, which is normal, versus a truck driven in a pedestrian area, which is abnormal. The reliance on context essentially generates an unbounded set of possible anomaly types. Coupled with the difficulty of collecting sufficient data for certain anomaly types (it is not ethical to fight or hurt people just to obtain video examples), the reliance on context makes it nearly impossible to gather abnormal training data. 

In the recent literature, we identified two distinct formulations to deal with the difficulty of the video anomaly detection task. On the one hand, we have the mainstream formulation, adopted in works such as \cite{Antic-ICCV-2011,Cheng-CVPR-2015,Dong-Access-2020,Hasan-CVPR-2016,Ionescu-WACV-2019,Kim-CVPR-2009,Lee-TIP-2019,Li-PAMI-2014,Liu-CVPR-2018,Lu-ICCV-2013,Luo-ICCV-2017,Mahadevan-CVPR-2010,Mehran-CVPR-2009,Park-CVPR-2020,Ramachandra-WACV-2020a,Ramachandra-WACV-2020b,Ravanbakhsh-WACV-2018,Ravanbakhsh-ICIP-2017,Sabokrou-IP-2017,Tang-PRL-2020,Wu-TNNLS-2019,Xu-CVIU-2017,Zhao-CVPR-2011,Zhang-PR-2020, Dutta-AAAI-2015, Zhang-PR-2016, Cong-CVPR-2011, Ren-BMVC-2015}, treating anomaly detection as a one-class classification (or outlier detection) task. In this formulation, training videos contain only normal events, while test videos encompass both normal and abnormal events. Under this scenario, methods learn a model of normality from familiar events, labeling unfamiliar events as abnormal at inference time. While framing anomaly detection as an outlier detection task preserves the open-set characteristic of anomaly types, the models proposed under this formulation usually obtain lower performance rates, as they lack knowledge of abnormal examples. On the other hand, we have the alternative formulation, considered in works such as \cite{Purwanto-ICCV-2021,Zaheer-ECCV-2020,Zhong-CVPR-2019,Feng-CVPR-2021,Tian-ICCV-2021}, treating anomaly detection as a weakly-supervised action recognition task, where training videos contain both normal and abnormal events, but the annotation is provided at the video level. This formulation corresponds to a closed-set evaluation scenario where training and test anomalies belong to the same action categories, failing to test the capability of systems at detecting unseen anomaly types.

\begin{figure}[!t]
\begin{center}
\centerline{\includegraphics[width=1.0\linewidth]{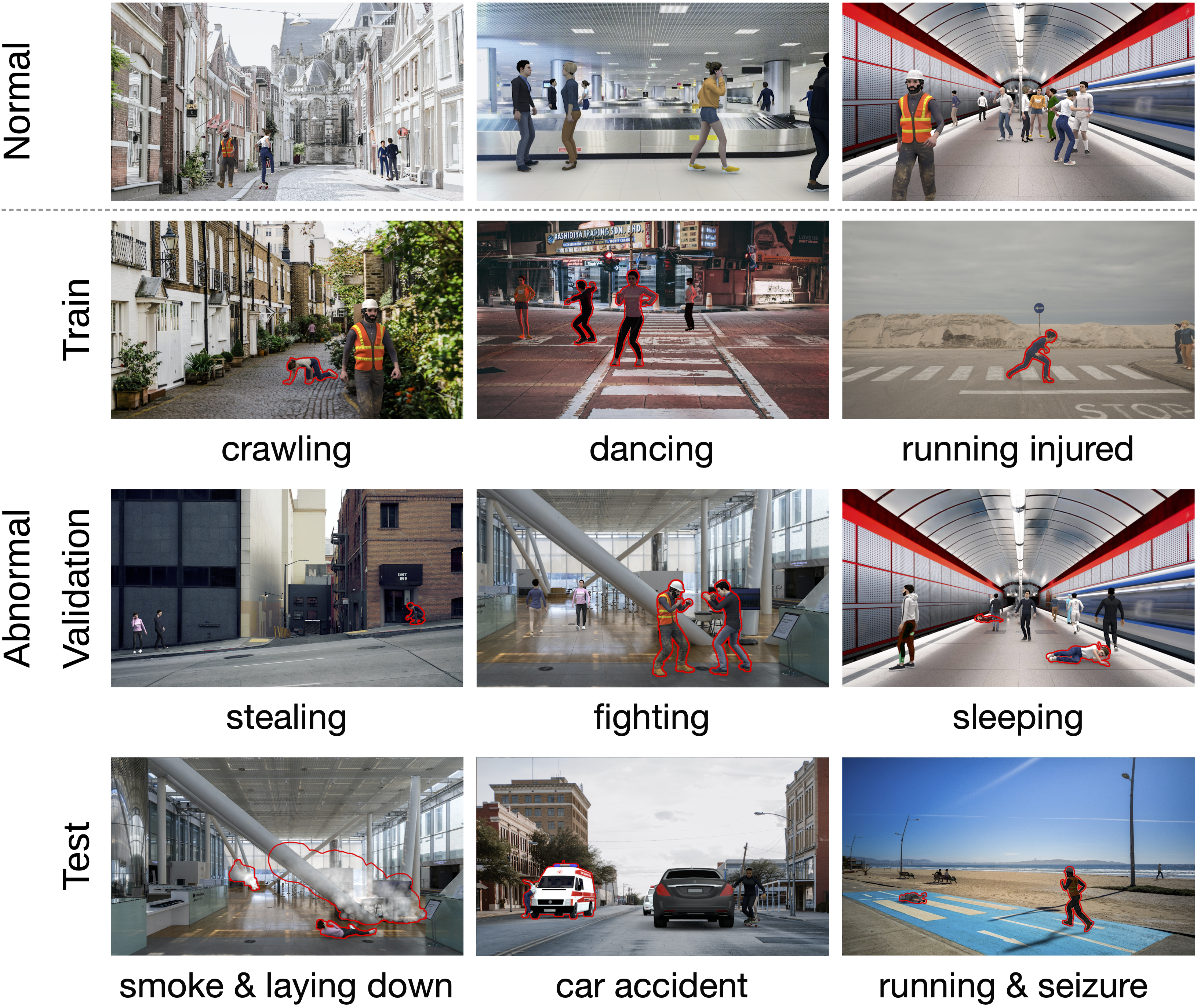}}
\vspace{-0.25cm}
\caption{Normal and abnormal examples from various scenes included in our data set. Objects performing a variety of abnormal actions (\eg: \emph{fighting}, \emph{crawling}, \emph{dancing}, \emph{sleeping}, \emph{running}) 
are emphasized through a red contour. Notice that the abnormal actions included in the test set belong to distinct categories from the abnormal actions used for training and validation, leading to a supervised open-set setting. More examples are provided in the supplementary. Best viewed in color.}
\label{fig_ubnormal_examples}
\end{center}
\vspace{-0.9cm}
\end{figure}

To this end, we propose a novel formulation that frames video anomaly detection as a supervised open-set classification problem. In our formulation, both normal and abnormal events are available at training time, but the anomalies that occur at inference time belong to a distinct set of anomaly types (categories). The main advantages of posing anomaly detection as a supervised open-set problem are: $(i)$ enabling the use of fully-supervised models due to the availability of anomalies at training time, $(ii)$ enabling the evaluation of models under unexpected anomaly types due to the use of disjoint sets of anomaly categories at training and test time, and $(iii)$ enabling the fair comparison between one-class open-set methods and weakly-supervised closed-set methods. As for the one-class problem formulation, an issue with the supervised open-set formulation lies in the difficulty of collecting abnormal data from the real world. To alleviate this issue, we propose \emph{UBnormal}, a new benchmark comprising multiple virtual scenes for video anomaly detection. Our scenes are generated in Cinema4D using virtual animated characters and objects that are placed in real-world backgrounds, as shown in Figure~\ref{fig_ubnormal_examples}. To the best of our knowledge, this is the first data set for supervised open-set anomaly detection in video.

Although UBnormal enables both the development and training of models in ideal conditions (with full supervision), and the evaluation of models in adverse conditions (on unknown anomaly types), its simulated scenes belong to a different data distribution than natural scenes. Therefore, it might be unclear what is the behavior of fully-supervised models on real-world scenes. To this end, we propose to overcome the distribution gap using CycleGAN \cite{Zhu-ICCV-2017}, translating simulated objects from UBnormal to real-world benchmarks such as Avenue \cite{Lu-ICCV-2013} and ShanghaiTech \cite{Luo-ICCV-2017}. We then employ a state-of-the-art multi-task learning framework \cite{Georgescu-CVPR-2021} for anomaly detection, introducing a new proxy task to discriminate between normal and abnormal samples from our data set. Our results show that UBnormal can enhance the performance of the state-of-the-art multi-task learning framework on both data sets. Interestingly, we provide empirical evidence indicating that performance gains can be achieved even without trying to close the distribution gap with CycleGAN. This shows that UBnormal can directly improve the performance of state-of-the-art models in the real-world.

In summary, our contribution is threefold:
\begin{itemize}
\vspace{-0.15cm}
    \item We pose video anomaly detection as a supervised open-set task, introducing UBnormal, a new data set of $29$ virtual scenes with $236,\!902$ video frames. \vspace{-0.25cm}
    \item We show that abnormal training videos are helpful to a variety of state-of-the-art models \cite{Georgescu-TPAMI-2021,Sultani-CVPR-2018,Bertasius-ICML-2021} for abnormal event detection. \vspace{-0.25cm}
    \item We conduct a data-centric study, showing that UBnormal data can bring performance gains to a recent state-of-the-art method \cite{Georgescu-CVPR-2021} on two natural scene data sets, Avenue and ShanghaiTech.
\end{itemize}

\section{Related Work}

\subsection{Video Anomaly Detection Methods}

A large body of recent works \cite{Antic-ICCV-2011,Cheng-CVPR-2015,Dong-Access-2020,Hasan-CVPR-2016,Ionescu-WACV-2019,Kim-CVPR-2009,Lee-TIP-2019,Li-PAMI-2014,Liu-CVPR-2018,Lu-ICCV-2013,Luo-ICCV-2017,Mahadevan-CVPR-2010,Mehran-CVPR-2009,Park-CVPR-2020,Ramachandra-WACV-2020a,Ramachandra-WACV-2020b,Ravanbakhsh-WACV-2018,Ravanbakhsh-ICIP-2017,Sabokrou-IP-2017,Wu-TNNLS-2019,Xu-CVIU-2017,Zhao-CVPR-2011,Zhang-PR-2020, Dutta-AAAI-2015, Zhang-PR-2016, Cong-CVPR-2011, Ren-BMVC-2015, Madan-ICCVW-2021, Ramachandra-MVA-2021, Chang-RP-2022, Li-CVIU-2021, Yang-Access-2021,Szymanowicz-CVPRW-2021, Yu-TNNLS-2021,Georgescu-CVPR-2021,Georgescu-TPAMI-2021,Ionescu-CVPR-2019,Lee-TIP-2019,Liu-ICCV-2021,Tang-PRL-2020,Vu-AAAI-2019,Wang-ACMMM-2020,Yu-ACMMM-2020,Zaheer-CVPR-2020,Zaheer-ECCV-2020,Astrid-ICCVW-2021,Astrid-BMVC-2021} treats anomaly detection in video as a one-class classification (outlier detection) task, training models without having access to abnormal samples.
These works can be divided into three categories corresponding to the level at which the algorithm is applied: frame, patch or object. 
For instance, Yu \etal \cite{Yu-TNNLS-2021} proposed a frame-level framework which employs the adversarial learning of past and future events to detect anomalies, instead of requiring complementary information such as optical flow or explicit abnormal samples during training. 
Ramachandra \etal \cite{Ramachandra-MVA-2021} proposed a patch-level framework to localize anomalies in video by using a Siamese network to learn a metric from pairs of video patches. The learned metric is used to measure the perceptual distance between video patches from a test video and video patches from the normal training videos.
Georgescu \etal \cite{Georgescu-CVPR-2021} employed self-supervised multi-task learning at the object level to detect anomalies in video. The framework uses a total of four proxy tasks, three based on self-supervision and one based on knowledge distillation. The final anomaly score for each object detected in a frame is computed by averaging the anomaly scores predicted for each proxy task.

\begin{table*}[t]
\setlength\tabcolsep{2.0pt}
\small{
\begin{center}
\begin{tabular}{|l|r|r|r|r|r|r|c|c|c|c|}
 \hline 
    &   \multicolumn{6}{|c|}{$\#$frames}     &  & & & \\
\cline{2-7}
 {Data set} &  \multicolumn{1}{|c|}{total}  &   \multicolumn{1}{|c|}{training}    &   \multicolumn{1}{|c|}{validation}   &  \multicolumn{1}{|c|}{test}   &   \multicolumn{1}{|c|}{normal}   & \multicolumn{1}{|c|}{abnormal} &  \multicolumn{1}{|c|}{$\#$anomalies} &  \multicolumn{1}{|c|}{$\#$scenes} & $\#$anomaly  & open \\
 & & & & & & & & & \multicolumn{1}{|c|}{types} & set \\
\hline
\hline
CUHK Avenue\cite{Lu-ICCV-2013}    &  $30,\!652$   &  $15,\!328$  & 	-   &  $15,\!324$   &  $26,\!832$  & $3,\!820$ & $77^\ddag$ & $1$ & $5$  & \cmark \\
\hline
ShanghaiTech \cite{Luo-ICCV-2017} & $317,\!398$   &  $274,\!515$  & 	-   &  $42,\!883$   &    $\mathbf{300,\!308}$  & $17,\!090$ & $158^\dag$ & $13$ & $11$  & \cmark\\
\hline
Street Scene \cite{Ramachandra-WACV-2020a}  &  $203,\!257$  &  $56,\!847$  & 	-   &  $146,\!410$    &  $159,\!341$  & $43,\!916$ & $205$ & $1$ & $17$ & \cmark\\
\hline
Subway Entrance \cite{Adam-PAMI-2008} &  $144,\!250$ &  $76,\!453$  & 	-   &  $67,\!797$      &  $132,\!138^\dag$  & $12,\!112^\dag$ & $51^\dag$ & $1$ & $5$ &  \cmark\\
\hline
Subway Exit \cite{Adam-PAMI-2008}  &  $64,\!901$ &  $22,\!500$  & 	-   &  $42,\!401$      &  $60,\!410^\dag$  & $4,\!491^\dag$ & $14^\dag$ & $1$ & $3$ &  \cmark \\
\hline
UCF-Crime \cite{Sultani-CVPR-2018} &  $\mathbf{13,\!741,\!393}$  &  $\mathbf{12,\!631,\!211}$  & 	-   &  $\mathbf{1,\!110,\!182}$    &  NA  & NA & NA & NA & $13$ & \xmark \\
\hline
UCSD Ped1 \cite{Mahadevan-CVPR-2010} &  $14,\!000$   & $6,\!800$  & 	-   &  $7,\!200$   &    $9,\!995$  & $4,\!005$ & $61^\ddag$ & $1$ & $5$ &  \cmark \\
\hline
UCSD Ped2 \cite{Mahadevan-CVPR-2010} &  $4,\!560$   & $2,\!550$  & 	-   &  $2,\!010$   &   $2,\!924$  & $1,\!636$ & $21^\ddag$ & $1$ & $5$ &  \cmark \\
\hline
UMN \cite{Mehran-CVPR-2009} &  $7,\!741$ &  $ NA$  & 	-   &  $NA$      &  $6,\!165$  & $1,\!576$ & $11$ & $3$ & $1$ &  \cmark \\
\hline 
{\bf UBnormal} (ours)   &  $236,\!902$  & 	$116,\!087$   &  $\mathbf{28,\!175}$   &  $92,\!640$   &  $147,\!887$  & $\mathbf{89,\!015}$ & $\mathbf{660}$ & $\mathbf{29}$ & $\mathbf{22}$ &  \cmark \\
\hline
\end{tabular}
\end{center}
}
\vspace{-0.5cm}
\caption{Statistics about our novel benchmark versus existing anomaly detection data sets. Compared with other open-set benchmarks, we have a higher number of abnormal events from a broader set of action categories (anomaly types), occurring across a larger set of scenes. Top number in each column is in bold text. Legend: $\dag$ -- computed based on tracks from \cite{Georgescu-TPAMI-2021}; $\ddag$ -- computed based on tracks from \cite{Ramachandra-WACV-2020a}. 
\label{table:datasets}}  
\vspace{-0.3cm}
\end{table*}

Another body of works \cite{Purwanto-ICCV-2021,Sultani-CVPR-2018,Zaheer-ECCV-2020,Zhong-CVPR-2019,Feng-CVPR-2021,Tian-ICCV-2021} treats anomaly detection as a weakly-supervised action recognition task. In this line of work, an algorithm is trained on both normal and abnormal videos, but the abnormal videos are annotated at the video level. The anomalies occurring at test time belong to the same action categories as the training anomalies, leading to an easier closed-set problem. Sultani \etal \cite{Sultani-CVPR-2018} proposed an algorithm based on multiple instance learning, building a deep anomaly ranking model that predicts high anomaly scores for abnormal video segments. To increase the anomaly detection performance, Feng \etal \cite{Feng-CVPR-2021} proposed a multiple instance self-training framework consisting of a multiple instance pseudo-label generator and a self-guided attention encoder in order to focus on the anomalous regions in each frame. We notice that weakly-supervised anomaly detection frameworks \cite{Purwanto-ICCV-2021,Zaheer-ECCV-2020,Zhong-CVPR-2019,Feng-CVPR-2021,Tian-ICCV-2021} are often evaluated on the ShanghaiTech or UCSD Ped data sets, but there are no official splits for weakly-supervised training on these data sets. Due to the unavailability of official splits, researchers tend to use their own data splits, leading to unfair comparisons between methods. To this end, we emphasize that the comparison between weakly-supervised and outlier detection (one-class) frameworks is unfair. This is because the former methods gain knowledge from the abnormal training data which is not available for the latter methods. If the anomaly types in training and test would be disjoint, the comparison could become more fair.
 
The existing video anomaly detection methods either use normal training data only \cite{Antic-ICCV-2011,Cheng-CVPR-2015,Dong-Access-2020,Hasan-CVPR-2016,Ionescu-WACV-2019,Kim-CVPR-2009,Lee-TIP-2019,Li-PAMI-2014,Liu-CVPR-2018,Lu-ICCV-2013,Luo-ICCV-2017,Mahadevan-CVPR-2010,Mehran-CVPR-2009,Park-CVPR-2020,Ramachandra-WACV-2020a,Ramachandra-WACV-2020b,Ravanbakhsh-WACV-2018,Ravanbakhsh-ICIP-2017,Sabokrou-IP-2017,Wu-TNNLS-2019,Xu-CVIU-2017,Zhao-CVPR-2011,Zhang-PR-2020, Dutta-AAAI-2015, Zhang-PR-2016, Cong-CVPR-2011, Ren-BMVC-2015, Madan-ICCVW-2021, Ramachandra-MVA-2021, Chang-RP-2022, Li-CVIU-2021, Yang-Access-2021,Szymanowicz-CVPRW-2021, Yu-TNNLS-2021,Georgescu-CVPR-2021,Georgescu-TPAMI-2021,Ionescu-CVPR-2019,Lee-TIP-2019,Liu-ICCV-2021,Tang-PRL-2020,Vu-AAAI-2019,Wang-ACMMM-2020,Yu-ACMMM-2020,Zaheer-CVPR-2020,Zaheer-ECCV-2020,Astrid-BMVC-2021,Astrid-ICCVW-2021}, or abnormal training data with video-level annotations~\cite{Purwanto-ICCV-2021,Zaheer-ECCV-2020,Zhong-CVPR-2019,Feng-CVPR-2021,Purwanto-ICCV-2021,Tian-ICCV-2021}. To our knowledge, there is no off-the-shelf method that can be applied on UBnormal and take full advantage of its supervised open-set nature. To this end, we introduce minimal changes to the considered baselines to leverage the availability of abnormal training data.

\subsection{Video Anomaly Detection Data Sets}

To date, there are quite a few data sets available for anomaly detection in video. We report several statistics about the most utilized data sets in Table~\ref{table:datasets}. While there are several data sets that preserve the open-set characteristic of anomaly detection \cite{Lu-ICCV-2013, Luo-ICCV-2017,Ramachandra-WACV-2020a,Adam-PAMI-2008,Mahadevan-CVPR-2010,Mehran-CVPR-2009}, to the best of our knowledge, there is only one data set for closed-set anomaly detection, namely UCF-Crime \cite{Sultani-CVPR-2018}.

The open-set benchmarks can be separated into two categories according to the number of scenes. The CUHK Avenue \cite{Lu-ICCV-2013}, Street Scene \cite{Ramachandra-WACV-2020a}, Subway Entrance \cite{Adam-PAMI-2008}, Subway Exit \cite{Adam-PAMI-2008} and UCSD Ped \cite{Mahadevan-CVPR-2010} data sets form the category of single-scene benchmarks \cite{Ramachandra-PAMI-2020}, enabling the successful use of frame-level methods that learn very specific normality models (adapted to a particular scene). In contrast, ShanghaiTech \cite{Luo-ICCV-2017} and UMN \cite{Mehran-CVPR-2009} belong to the category of multiple-scene benchmarks, which tests the capability of methods to build more generic normality models, that perform well on several scenes. Street Scene \cite{Ramachandra-WACV-2020a} is the largest single-scene data set, containing $203,\!257$ frames. Despite being the largest data set for the single-scene scenario, it is not representative of real-world models that are expected to run across multiple scenes (as long as the normal behavior is similar across scenes). With $317,\!398$ frames, the largest multiple-scene data set for open-set anomaly detection is ShanghaiTech \cite{Luo-ICCV-2017}. Despite being the largest in its scenario, it has a small number of anomalies ($158$) belonging to only $11$ anomaly types. 

Most of the existing data sets contain anomalies related to human-interaction \cite{Lu-ICCV-2013,Adam-PAMI-2008,Mahadevan-CVPR-2010,Mehran-CVPR-2009} or vehicles in pedestrian areas, but the anomalies are usually staged and lack variety. For example, in UCSD Ped2 \cite{Mahadevan-CVPR-2010}, each anomalous event is related to bicycles, skateboards or cars in a pedestrian area. 
Similarly, Subway \cite{Adam-PAMI-2008} contains only anomalies related to people, such as people walking in the wrong direction or people jumping over the turnstiles. The size of Subway is quite large, with $144,\!250$ frames for the Entrance video and $64,\!901$ frames for the Exit video, but the number of anomaly types is very small, with only $5$ anomaly types for Entrance and $3$ anomaly types for Exit, respectively. 

The closed-set UCF-Crime benchmark introduced by Sultani \etal \cite{Sultani-CVPR-2018} contains $13$M frames from videos retrieved from YouTube and LiveLeak, using text queries. The data set contains $13$ anomaly categories that can be found in both training and test videos. 
UCF-Crime does not follow the open-set paradigm for anomaly detection in video, where the actions from the training set are different for those in the test set. Moreover, the data set does not contain pixel-level anomaly annotations.

To the best of our knowledge, we are the first to propose a benchmark for supervised open-set anomaly detection. We consider several factors that justify the need for a novel anomaly detection benchmark. First and foremost, unlike existing data sets, our benchmark contains anomalies with pixel-level annotations in the training set. The anomaly types from the training set are different from the anomaly types seen in the test set, conforming to the open-set constraint. Second, none of the existing data sets have a validation set, a mandatory requirement for many machine learning algorithms relying on hyperparameter tuning. This leaves two options, either tune the model on the test data, inherently overfitting the model to the test set, or refrain from hyperparameter tuning, likely leading to suboptimal results. In contrast to existing benchmarks, we are the first to provide a validation set. It contains anomalies belonging to a different set of action categories than the set of action categories available at test time. This ensures the possibility to perform model tuning without overfitting to the test set. Third, some of the existing data sets, \eg UCSD Ped \cite{Mahadevan-CVPR-2010} and UMN \cite{Mehran-CVPR-2009}, are already saturated (the performance surpasses $99\%$ in terms of the frame-level AUC), while the performance on other benchmarks, \eg Avenue \cite{Lu-ICCV-2013} and Subway \cite{Adam-PAMI-2008}, surpasses $90\%$ in terms of the frame-level AUC. For example, the results reported in \cite{Georgescu-TPAMI-2021} on Avenue (a micro AUC of $92.3\%$) and UCSD Ped2 (a micro AUC of $98.7\%$) are significantly higher compared to the results obtained by the same method on the UBnormal data set (a micro AUC of $59.3\%$). In general, our experiments show that UBnormal is significantly more challenging, likely due to the higher variation of anomaly types and scenes. Considering all these aspects, we believe that UBnormal will likely contribute towards the development of future anomaly detection models.

\subsection{Open-Set Action Recognition}

With respect to the definition of open-set video recognition of Geng \etal \cite{Geng-PAMI-2020}, in our setting, the normal classes are known known classes (KKC) and the abnormal classes used at test time are unknown unknown classes (UUC). We consider the abnormal actions used at training time as known unknown classes (KUC). However, KUC samples will not appear at test time. To this end, our setting can be considered as \emph{supervised open-set}, which is distinct from the classical \emph{open-set} setting. Moreover, we underline that the anomaly detection task is distinct from action recognition, \ie abnormal actions need to be detected and localized in long videos. One video may contain multiple normal and abnormal events happening at the same time. Hence, a pure action recognition approach is not likely to provide optimal results in anomaly detection. Hence, we consider works on open-set action recognition \cite{Bao-ICCV-2021,Chen-PAMI-2021} as very distantly related.

\section{UBnormal Benchmark}


\noindent
\textbf{Scenes.}
The UBnormal benchmark is generated using the Cinema4D software, which allows us to create scenes using 2D background images and 3D animations. We select a total of $29$ natural images that represent street scenes, train stations, office rooms, among others. In the selected background images, we make sure to eliminate people, cars or other objects that should belong to the foreground. From each natural image, we create a virtual 3D \emph{scene} and generate (on average) $19$ videos per scene. For each scene, we generate both normal and abnormal videos. The proportion of normal versus abnormal videos in the entire UBnormal data set is close to $1\!:\!1$.

\noindent
\textbf{Action categories.}
We consider the following events as normal, for all our video scenes: \emph{walking}, \emph{talking on the phone}, \emph{walking while texting}, \emph{standing}, \emph{sitting}, \emph{yelling} and \emph{talking with others}. In addition, we introduce a total of $22$ abnormal event types, as follows: \emph{running}, \emph{falling}, \emph{fighting}, \emph{sleeping}, \emph{crawling}, \emph{having a seizure}, \emph{laying down}, \emph{dancing}, \emph{stealing}, \emph{rotating 360 degrees}, \emph{shuffling}, \emph{walking injured}, \emph{walking drunk}, \emph{stumbling walk}, \emph{people and car accident}, \emph{car crash}, \emph{running injured}, \emph{fire}, \emph{smoke}, \emph{jaywalking}, \emph{driving outside lane} and \emph{jumping}.
We organize the anomalous event types such that those included in the test set are different from those found in the training and validation sets. Hence, the test set includes the following anomalies: \emph{running}, \emph{having a seizure}, \emph{laying down}, \emph{shuffling}, \emph{walking drunk}, \emph{people and car accident}, \emph{car crash}, \emph{jumping}, \emph{fire}, \emph{smoke}, \emph{jaywalking} and \emph{driving outside lane}. The following anomalies are included in the training set: \emph{falling}, \emph{dancing}, \emph{walking injured}, \emph{running injured}, \emph{crawling} and \emph{stumbling walk}. The rest of the anomalies are added to the validation set. 

\noindent
\textbf{Variety.}
In order to increase the variety of our data set, we include multiple object categories, such as people, cars, skateboards, bicycles and motorcycles. Unlike other data sets (CUHK Avenue\cite{Lu-ICCV-2013}, ShanghaiTech \cite{Luo-ICCV-2017}, UCSD Ped \cite{Mahadevan-CVPR-2010}), these objects can perform both normal and abnormal actions, thus being present in both normal and abnormal videos. Hence, simply labeling an object as abnormal because it belongs to an unseen category is no longer possible. To further increase the diversity of the data set, we include foggy scenes, night scenes, and fire and smoke as abnormal events.
In existing data sets, a single person or a small group of people perform most of the abnormal actions. We can take the person with blue pants that is running or throwing a backpack or some papers in CUHK Avenue \cite{Lu-ICCV-2013} as example, or the group of guys that are running and fighting in various scenes from ShanghaiTech \cite{Luo-ICCV-2017}. Different from such data sets, the anomalous events in the UBnormal data set are performed by various characters. We employ $19$ different characters to animate the videos. We also change the colors of their clothes or their hair color, increasing the diversity of the animated characters included in our benchmark. 
We provide several figures showcasing the variety of scenes, characters and actions of UBnormal in the supplementary.

\noindent
\textbf{Data generation and annotation.}
The anomalies in the UBnormal data set are annotated at the pixel level. For each synthetic object (normal or abnormal) in the data set, we provide the segmentation mask and the object label (\emph{person}, \emph{car}, \emph{bicycle}, \emph{motorcycle} or \emph{skateboard}). In the process of simulating the events and generating the benchmark, we involved a team of six people for a period of three months. We generate all videos at $30$ FPS with the minimum height of a frame set to $720$ pixels. It takes about $15$ seconds to render one frame with the Cinema4D software, taking a total of $987$ hours ($41.1$ days) to render the entire data set. After generating the videos, four of our team members checked each generated video for incorrect occlusions, gravity-related issues or other visual inconsistencies, ensuring the high quality of the generated data set.

\section{Methods}

\noindent
\textbf{One-class open-set model.} As the first baseline for the UBnormal data set, we employ the state-of-the-art background-agnostic framework introduced in \cite{Georgescu-TPAMI-2021}. This is an object-level method which treats anomaly detection as a one-class classification task. The framework is composed of three auto-encoders and three classifiers. To increase the performance of the anomaly detection method, Georgescu \etal\cite{Georgescu-TPAMI-2021} proposed an adversarial learning scheme for the auto-encoders. To overcome the lack of abnormal samples during training, they created a scene-agnostic set of pseudo-anomalies. The pseudo-abnormal samples are used in the adversarial training process as adversarial examples and, in the training of the binary classifiers, as abnormal samples. 

In our first experiment, we employ the framework \cite{Georgescu-TPAMI-2021} without any modification as a baseline for our benchmark. Then, we add the abnormal samples from our training data set to the pool of pseudo-abnormal samples.

\noindent
\textbf{Supervised closed-set model.} As another baseline, we consider the supervised closed-set model proposed by Sultani \etal\cite{Sultani-CVPR-2018}. In this framework\cite{Sultani-CVPR-2018}, the normal and abnormal videos are represented as bags, and the video segments are instances in multiple instance learning. Sultani \etal\cite{Sultani-CVPR-2018} represented each video using features extracted by a pre-trained C3D~\cite{Tran-ICCV-2015} model. Using the video feature representation, they trained a feed-forward neural network such that the maximum score of the instances from the anomalous bag is higher than the maximum score of the instances from the normal bag. They also incorporated sparsity and smoothness constraints in the loss function to further boost the performance of the framework.
 
\noindent
\textbf{Action recognition framework.} The third baseline for our benchmark is an action recognition model. We choose the state-of-the-art model proposed by Bertasius \etal \cite{Bertasius-ICML-2021}. The TimeSformer~\cite{Bertasius-ICML-2021} architecture adjusts the standard transformer architecture to the video domain, by learning spatio-temporal features from sequences of frame patches. TimeSformer~\cite{Bertasius-ICML-2021} uses divided attention, \ie the spatial and temporal attention maps are learned separately. We train the TimeSformer model to distinguish between normal and abnormal actions. The model predicts the probability of a sequence of frames of being abnormal.

\noindent
\textbf{Self-supervised multi-task model.}
We employ the state-of-the-art multi-task learning framework of Georgescu \etal \cite{Georgescu-CVPR-2021} to show that UBnormal can be used to boost performance on two prominent anomaly detection data sets, namely CUHK Avenue \cite{Lu-ICCV-2013} and ShanghaiTech \cite{Luo-ICCV-2017}.  We hereby state that the main goal for conducting experiments with the state-of-the-art method of Georgescu \etal \cite{Georgescu-CVPR-2021} is to determine if data from the UBnormal data set can help to increase performance on two real-world data sets, regardless of how our data is integrated into the training sets of the respective real-world benchmarks. While we expect higher performance gains after performing domain adaption with CycleGAN (as detailed below), we emphasize that these gains would not be possible without the examples coming from our data set. 

The object-level method of Georgescu \etal \cite{Georgescu-CVPR-2021} is based on learning a single 3D convolutional neural network (CNN) on four proxy tasks, namely the arrow of time, motion irregularity, middle box prediction and model distillation. 
Further, we integrate the fifth proxy task ($T_5$) to discriminate between normal and abnormal objects from the UBnormal data set. Here, we consider two options: $(i)$ train the model directly on UBnormal examples, and $(ii)$ pass the objects seen only at training time through CycleGAN \cite{Zhu-ICCV-2017}, before training on the fifth proxy task. During inference, we expect the normal samples from the test set to be classified as normal, and the abnormal samples to be classified as abnormal. For each object, we create an object-centric temporal sequence by cropping the object bounding box from the frames $\{i-t,...,i-1, i, i+1,...,i+t\}$, following \cite{Georgescu-CVPR-2021}. The normal object-centric sequences are labeled with class $1$, while the abnormal sequences are labeled with class $2$. 

Let $f$ be the shared 3D CNN and $h_{T_5}$ be our abnormality head. Let $X^{(T_5)}$ be a normal or abnormal object-centric sequence of size $(2\cdot t+1) \times 64 \times 64 \times 3$.  We employ the cross-entropy loss to train the abnormality head:
\begin{equation}\label{eq_arrow_of_time}
\mathcal{L}_{T_5} \left( X^{(T_5)}, Y^{(T_5)}\right) = -\sum_{k=1}^{2} Y^{(T_5)}_k log\left(\hat{Y}^{(T_5)}_k\right) ,
\end{equation}
where $\hat{Y}^{(T_5)}\!=\!\mbox{softmax}\!\left( h_{T_5}\left(f(X^{(T_5)}\right) \right)$ and $Y^{(T_5)}$ is the one-hot encoding of the ground-truth label for $X^{(T_5)}$.

After integrating the fifth proxy task, the shared 3D CNN is trained with the following joint loss:
\begin{equation}\label{eq_total_loss}
\mathcal{L}_{\mbox{\scriptsize{total}}} = \mathcal{L}_{T_1} + \mathcal{L}_{T_2} + \mathcal{L}_{T_3} + \lambda \cdot \mathcal{L}_{T_4} + \mathcal{L}_{T_5}.
\end{equation}
Further details about $\mathcal{L}_{T_1}, \mathcal{L}_{T_2}, \mathcal{L}_{T_3}$ and $\mathcal{L}_{T_4}$ are provided in \cite{Georgescu-CVPR-2021}.
At inference time, the anomaly score for each object is computed using the following equation:
\begin{equation}\label{eq_total}
\begin{split}
\!\!\!\mbox{score}(X) &\!=\! \frac{1}{5}\Big(\hat{Y}^{(T_1)}_2 \!+ \hat{Y}^{(T_2)}_2 \!+ \mbox{avg}\!\left(\left|Y^{(T_3)}\!-\!\hat{Y}^{(T_3)} \right|\right) \\
& + avg\!\left(\left|Y^{(T_4)}_{\mbox{\scriptsize{YOLO}}}\!-\! \hat{Y}^{(T_4)}_{\mbox{\scriptsize{YOLO}}} \right| \right ) + \hat{Y}^{(T_5)}_2 \Big) .
\end{split}
\end{equation}
The notations $\hat{Y}^{(T_1)}_2, \hat{Y}^{(T_2)}_2, Y^{(T_3)}, \hat{Y}^{(T_3)}, Y^{(T_4)}_{\mbox{\scriptsize{YOLO}}}$ and $\hat{Y}^{(T_4)}_{\mbox{\scriptsize{YOLO}}}$ are defined in \cite{Georgescu-CVPR-2021}.

\section{Experiments} 
\subsection{Setup and Implementation Details}

\noindent
\textbf{Data sets.} Aside from reporting results on UBnormal, we evaluate how helpful the data from UBnormal is for other real-world anomaly detection benchmarks. To this end, we consider the popular CHUK Avenue \cite{Lu-ICCV-2013} and ShanghaiTech \cite{Luo-ICCV-2017} data sets. We provide more details about these benchmarks in Table~\ref{table:datasets}.

\noindent
\textbf{Evaluation measures.}
As evaluation metrics, we consider the widely-used area under the curve (AUC), computed with respect to the ground-truth frame-level annotations, as well as the region-based detection criterion (RBDC) and track-based detection criterion (TBDC) introduced by Ramachandra \etal\cite{Ramachandra-WACV-2020a}. For the frame-level AUC, we consider both micro and macro versions, following \cite{Georgescu-TPAMI-2021}. 

\noindent
\textbf{Learning and parameter tuning.} In order to train the one-class open-set model and the supervised closed-set framework on UBnormal, we use the official code provided by Georgescu \etal\cite{Georgescu-TPAMI-2021} and Sultani \etal\cite{Sultani-CVPR-2018}, respectively. We train the models following the instructions given by the authors, without changing any of the hyperparameters. 

To train the TimeSformer model on UBnormal, we rely on the official implementation released by Bertasius \etal \cite{Bertasius-ICML-2021}. We use the default version of TimeSformer, which processes $8$ frames, each having a spatial resolution of $224\times224$ pixels. In order to adapt the TimeSformer model to abnormal event detection, we change the last fully-connected layer to distinguish between two classes, \emph{normal} versus \emph{abnormal}. We start from the pre-trained TimeSformer and fine-tune it for $20$ epochs on the UBnormal benchmark, with the learning rate set to $5 \cdot 10^{-4}$. We perform experiments with the default video sample rate ($1/32$), as well as two additional sample rates ($1/8$ and $1/4$) that seemed more suitable for UBnormal, at least from our perspective.

\begin{figure}[!t]
\begin{center}
\centerline{\includegraphics[width=0.9\linewidth]{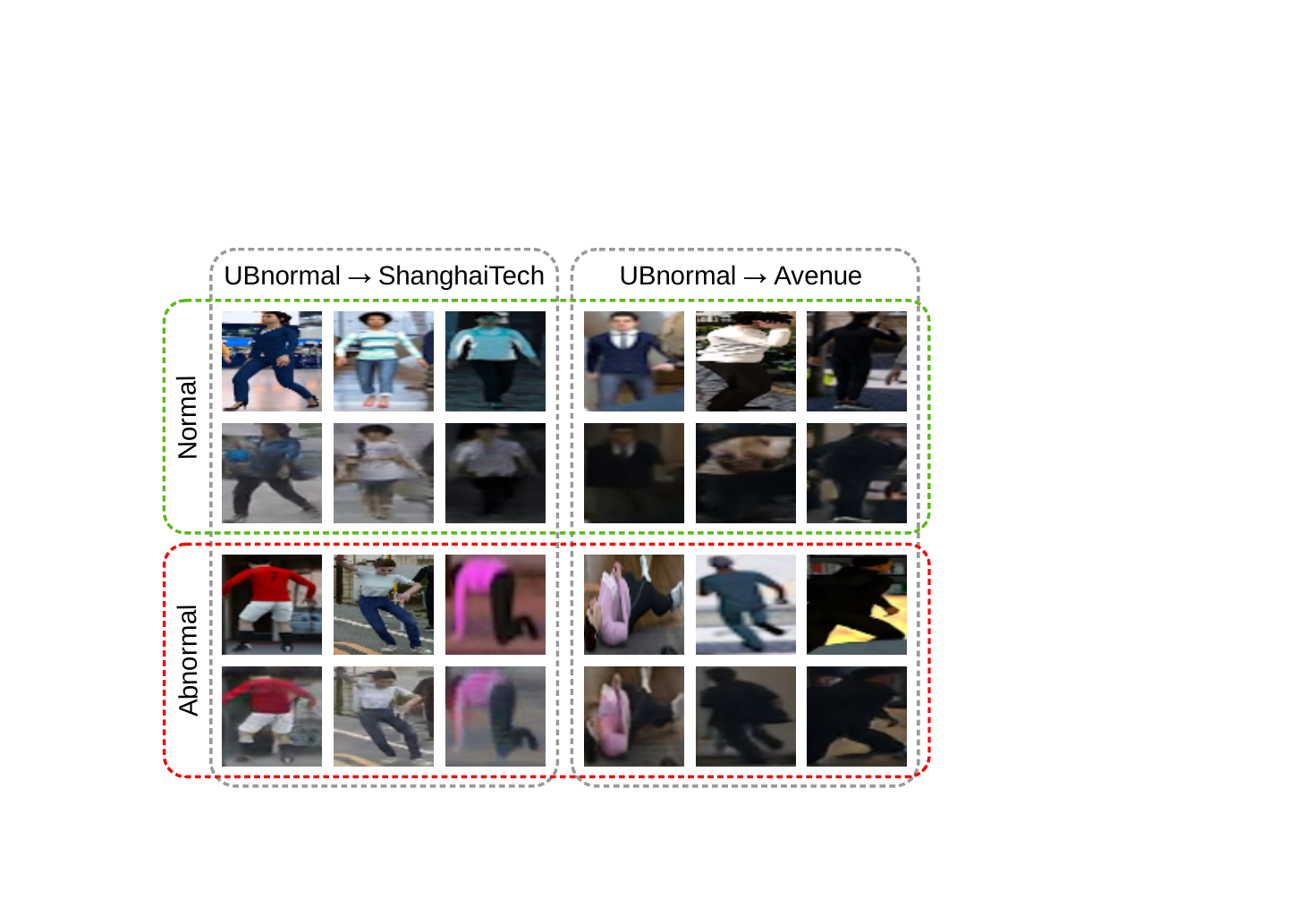}}
\vspace{-0.2cm}
\caption{Normal and abnormal objects from UBnormal (above) and the corresponding CycleGAN translations to ShanghaiTech and Avenue (below), respectively. Best viewed in color.}
\label{fig:cyclegan}
\end{center}
\vspace{-0.8cm}
\end{figure}

For the experiments on Avenue and ShanghaiTech,
we use the official code and the hyperparameters reported by Georgescu \etal \cite{Georgescu-CVPR-2021} to train the self-supervised multi-task model. To adapt UBnormal examples to Avenue and ShanghaiTech, respectively, we train CycleGAN \cite{Zhu-ICCV-2017} using the official code with the default hyperparameters. We apply CycleGAN at the object level. In order to extract the object bounding boxes from the UBnormal training set, we rely on the ground-truth segmentation masks. To detect the objects in Avenue and ShanghaiTech, we apply the pre-trained YOLOv3 \cite{Redmon-arXiv-2018} object detector, following \cite{Georgescu-CVPR-2021}. We train the CycleGAN model only on the training data from UBnormal and Avenue or ShanghaiTech, respectively. We optimize the CycleGAN model for $10$ epochs on each of the two data set pairs. In Figure~\ref{fig:cyclegan}, we present a few samples of the translated objects, before and after applying CycleGAN.

\begin{table*}[t]
\setlength\tabcolsep{2.0pt}
\small{
\begin{center}
\begin{tabular}{|l|c|c|c|c|c|c|c|c|}
 \hline 
                                                          &    \multicolumn{4}{|c|}{{Validation}}  &    \multicolumn{4}{|c|}{{Test}} \\ 
\cline{2-9}
Method       &    \multicolumn{2}{|c|}{{AUC}}   & RBDC   & TBDC &    \multicolumn{2}{|c|}{{AUC}}   & RBDC   & TBDC\\
\cline{2-3}
\cline{6-7}                                                                &    Micro      &  Macro           &          &      &    Micro      &  Macro           &          & \\ 
\hline 
\hline
Georgescu \etal\cite{Georgescu-TPAMI-2021}                      & $58.5$ & $94.4$ &  $18.580$ &  $48.213$  &  $59.3$     &   $84.9$         & $21.907$  & $53.438$\\
Georgescu \etal\cite{Georgescu-TPAMI-2021} + UBnormal anomalies & $68.2$ & $\mathbf{95.3}$ &  $\mathbf{28.654}$ &  $\mathbf{58.097}$  &  $61.3$     &   $\mathbf{85.6}$         & $\mathbf{25.430}$  & $\mathbf{56.272}$\\
\hline 
Sultani \etal\cite{Sultani-CVPR-2018}  (pre-trained)             &   $61.1$ & $89.4$ & $0.001$ & $0.012$ & $49.5$      &   $77.4$         & $0.001$  & $0.001$\\
Sultani \etal\cite{Sultani-CVPR-2018}  (fine-tuned)              &   $51.8$ & $88.0$ & $0.001$ & $0.001$ & $50.3$      &   $76.8$         & $0.002$  & $0.001$\\
\hline 
Bertasius \etal\cite{Bertasius-ICML-2021} ($1/32$ sample rate, fine-tuned)              &   $\mathbf{86.1}$ & $89.2$ & $0.008$ & $0.021$&   $\mathbf{68.5}$      &   $80.3$         & $0.041$  & $0.053$\\
Bertasius \etal\cite{Bertasius-ICML-2021} ($1/8$ sample rate, fine-tuned)               &   $83.4$ & $90.6$ & $0.009$ & $0.023$&   $64.1$      &   $75.4$         & $0.040$  & $0.050$\\
Bertasius \etal\cite{Bertasius-ICML-2021} ($1/4$ sample rate, fine-tuned)               &   $78.5$ & $89.2$ & $0.006$ & $0.018$&   $61.9$      &   $75.4$         & $0.040$  & $0.057$\\
\hline 
\end{tabular}
\end{center}
} 
\vspace{-0.5cm}
\caption{Micro-averaged frame-level AUC, macro-averaged frame-level AUC, RBDC and TBDC scores (in \%) of the proposed baselines~\cite{Georgescu-TPAMI-2021,Sultani-CVPR-2018,Bertasius-ICML-2021} on the UBnormal data set. Although only one method \cite{Georgescu-TPAMI-2021} can perform anomaly localization, we report RBDC and TBDC scores for all baselines, for completeness. Best results are highlighted in bold. \label{table:results_ubnormal}}
\vspace{-0.3cm}
\end{table*}
 
\subsection{Anomaly Detection Results}

\begin{figure}[!t]
\begin{center}
\centerline{\includegraphics[width=1.0\linewidth]{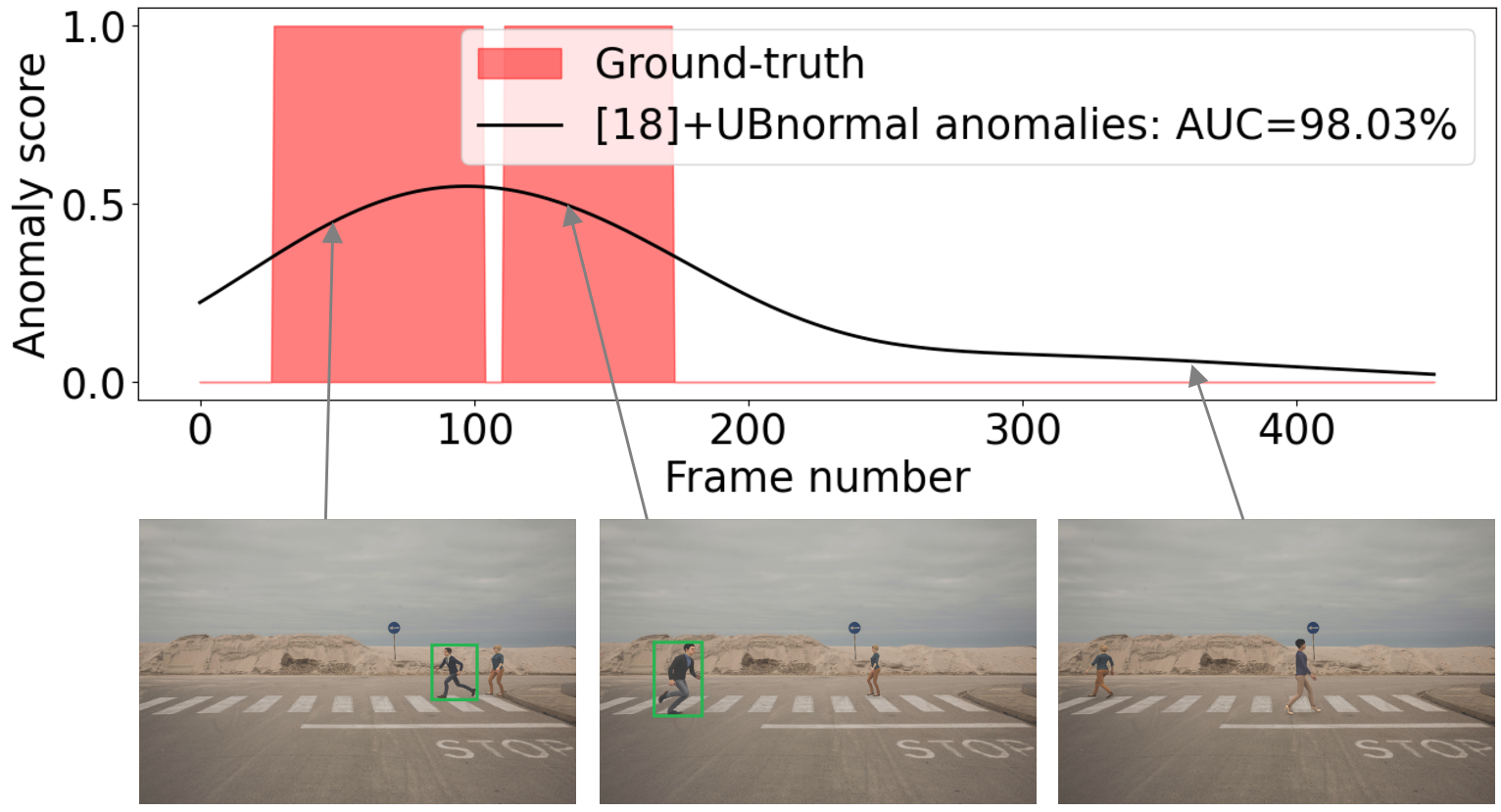}}
\vspace{-0.2cm}
\caption{Frame-level scores and anomaly localization examples for a test video from the UBnormal data set. Best viewed in color.}
\label{ubnormal_pami_abnormal_scene_11_scenario_2}
\end{center}
\vspace{-0.8cm}
\end{figure}

\noindent
\textbf{UBnormal.} 
In Table \ref{table:results_ubnormal}, we report the results obtained by the baselines~\cite{Georgescu-TPAMI-2021,Sultani-CVPR-2018, Bertasius-ICML-2021} on the validation and test sets of our benchmark. The method of Georgescu \etal \cite{Georgescu-TPAMI-2021} obtains a performance of $58.5\%$ in terms of the micro-averaged frame-level AUC on the validation set, while scoring $59.3\%$ on the test set. When we add the abnormal samples from our training data set to the pool of pseudo-abnormal samples, we observe performance gains for all four metrics. 
The framework of Georgescu \etal \cite{Georgescu-TPAMI-2021} is the only baseline method that performs anomaly localization, thus obtaining significantly higher RBDC and TBDC scores than the other two approaches. Nonetheless, we report the  RBDC and TBDC scores for all three baselines for completeness, but we acknowledge that two of the methods \cite{Sultani-CVPR-2018, Bertasius-ICML-2021} are only suitable for anomaly detection. Hence, in Figure~\ref{ubnormal_pami_abnormal_scene_11_scenario_2}, we present the frame-level anomaly scores and a set of anomaly localization examples provided by the framework of Georgescu \etal \cite{Georgescu-TPAMI-2021} for a test video from UBnormal. The illustrated anomalies represent \emph{people running}. More qualitative results are shown in the annotated videos uploaded at \small{\url{https://github.com/lilygeorgescu/UBnormal}}.

When employing the pre-trained network of Sultani \etal \cite{Sultani-CVPR-2018}, we obtain a micro-averaged frame-level AUC of $49.5\%$ on the test set. Further fine-tuning the network on the UBnormal data set increases the micro-averaged frame-level AUC to $50.3\%$ on the test set. The TimeSformer model \cite{Bertasius-ICML-2021} attains the highest micro-averaged frame-level AUC on both validation and test sets. Using the default video sample rate of $1/32$, we obtain a micro-averaged frame-level AUC of $68.5\%$ on the test set. As we thought that the video sample rate of $1/32$ was too high for anomaly detection, we also tried smaller sample rates ($1/8$, $1/4$), without prevail. Indeed, the empirical results show that the video sample rate of $1/32$ is optimal for TimeSformer.

In summary, we find that the TimeSformer model \cite{Bertasius-ICML-2021} is the method of choice for anomaly detection on UBnormal, while the framework of Georgescu \etal \cite{Georgescu-TPAMI-2021} remains the best choice when it comes to anomaly localization.


\begin{table}[t]
\setlength\tabcolsep{2.2pt}
\small{
\begin{center}
\begin{tabular}{|c|l|c|c|c|c|}
\hline %
\multirow{3}{*}{\rotatebox{90}{Year}} & \multirow{3}{*}{Method} &  \multicolumn{2}{|c|}{Avenue}  &  \multicolumn{2}{|c|}{ShanghaiTech}\\ 
\cline{3-6}
& & \multicolumn{2}{|c|}{AUC}          & \multicolumn{2}{|c|}{AUC}  \\
\cline{3-6}
& & Micro  & Macro         &    Micro  & Macro \\
\hline
\hline
\multirow{4}{*}{\rotatebox{90}{2018}}&Lee \etal~\cite{Lee-ICASSP-2018} & $87.2$ & -  & -  & - \\
&Liu \etal~\cite{Liu-CVPR-2018} & $85.1$ & $81.7^*$ &  $72.8$ & $80.6^*$ \\
&Liu \etal~\cite{Liu-BMVC-2018} & $84.4$ & - & -  & - \\
&Sultani \etal~\cite{Sultani-CVPR-2018} & - & - & - & $76.5$ \\
\hline
\multirow{7}{*}{\rotatebox{90}{2019}}&Gong \etal~\cite{Gong-ICCV-2019} & $83.3$  & - & $71.2$  & - \\
&Ionescu \etal~\cite{Ionescu-CVPR-2019} & $87.4^*$ & $90.4$ & $78.7^*$ & $84.9$ \\
&Ionescu \etal~\cite{Ionescu-WACV-2019} & $88.9$ & - & - & -  \\
&Lee \etal~\cite{Lee-TIP-2019} & $90.0$ & - & $76.2$  & - \\
&Nguyen \etal~\cite{Nguyen-ICCV-2019} & $86.9$  & - & - & - \\
&Vu \etal~\cite{Vu-AAAI-2019} & $71.5$ & - & - & - \\
&Wu \etal~\cite{Wu-TNNLS-2019} & $86.6$  & - & - & - \\
\hline
\multirow{11}{*}{\rotatebox{90}{2020}}
&Dong \etal~\cite{Dong-Access-2020} & $84.9$  & - & $73.7$  & - \\
&Doshi \etal~\cite{Doshi-CVPRW-2020a,Doshi-CVPRW-2020b} & $86.4$ & - & $71.6$ & - \\
&Ji \etal~\cite{Ji-IJCNN-2020} & $78.3$ & - & - & - \\
&Lu \etal~\cite{Lu-ECCV-2020} & $85.8$ & - & $77.9$  & - \\
&Park \etal~\cite{Park-CVPR-2020} & $88.5$ & - & $70.5$  & - \\
&Ramachandra \etal~\cite{Ramachandra-WACV-2020a} & $72.0$  & - & -  & - \\
&Ramachandra \etal~\cite{Ramachandra-WACV-2020b} & $87.2$  & - & - & - \\
&Sun \etal~\cite{Sun-ACMMM-2020} & $89.6$  & - & $74.7$ & - \\
&Tang \etal~\cite{Tang-PRL-2020} & $85.1$ & - & $73.0$  & - \\
&Wang \etal~\cite{Wang-ACMMM-2020} & $87.0$  & - & $79.3$ & - \\
&Yu \etal~\cite{Yu-ACMMM-2020} & $89.6$ & - & $74.8$  & - \\ 
\hline 
\multirow{10}{*}{\rotatebox{90}{2021}}
&Astrid \etal~\cite{Astrid-BMVC-2021} & $84.9$ & - & $76.0$ & - \\
&Astrid \etal~\cite{Astrid-ICCVW-2021} & $87.1$ & - & $73.7$ & - \\
&Chang \etal~\cite{Chang-RP-2022} &  $87.1$ & - & $73.7$ & - \\
&Georgescu  \etal~\cite{Georgescu-TPAMI-2021} & $92.3$ & $90.4$ & $82.7$ &  $89.3$ \\  
&Madan  \etal~\cite{Madan-ICCVW-2021} & $88.6$ & - & $74.6$ &  - \\  
&Li \etal ~\cite{Li-CVIU-2021} & $88.8$ & - & $73.9$ & - \\ 
&Liu \etal ~\cite{Liu-ICCV-2021} & $91.1$ & -  & $76.2$ & - \\
&Szymanowicz \etal ~\cite{Szymanowicz-CVPRW-2021} & $75.3$ & - & $70.4$ & - \\
&Yang \etal ~\cite{Yang-Access-2021} & $88.6$ & - & $74.5$ & - \\
&Yu \etal ~\cite{Yu-TNNLS-2021} & $90.2$ & -  & -  & - \\
\hline
&\cite{Georgescu-CVPR-2021}  & $91.5$ & $91.9$ & $82.4$ &  $89.3$ \\   
&\cite{Georgescu-CVPR-2021} + $T_5$ (ours) & $91.9$ & $92.4$ & $83.0$ &  $89.5$ \\  
&\cite{Georgescu-CVPR-2021} + $T_5$ + CycleGAN (ours) & $\mathbf{93.0}$ & $\mathbf{93.2}$ & $\mathbf{83.7}$ &  $\mathbf{90.5}$ \\  
\hline
\end{tabular}
\end{center}
} 
\vspace{-0.5cm}
\caption{Micro-averaged and macro-averaged frame-level AUC scores (in \%) of the state-of-the-art methods~\cite{Lee-ICASSP-2018,Liu-CVPR-2018,Liu-BMVC-2018,Sultani-CVPR-2018,Gong-ICCV-2019,Ionescu-CVPR-2019,Ionescu-WACV-2019,Lee-TIP-2019,Nguyen-ICCV-2019,Vu-AAAI-2019,Wu-TNNLS-2019,Dong-Access-2020,Doshi-CVPRW-2020a,Doshi-CVPRW-2020b,Ji-IJCNN-2020,Lu-ECCV-2020,Park-CVPR-2020,Ramachandra-WACV-2020a,Ramachandra-WACV-2020b,Sun-ACMMM-2020,Tang-PRL-2020,Wang-ACMMM-2020,Astrid-BMVC-2021,Astrid-ICCVW-2021,Chang-RP-2022,Georgescu-TPAMI-2021,Madan-ICCVW-2021,Li-CVIU-2021,Liu-ICCV-2021,Szymanowicz-CVPRW-2021,Yang-Access-2021,Yu-TNNLS-2021} versus an approach based on the framework proposed in \cite{Georgescu-CVPR-2021}, enhanced by us with the fifth proxy task ($T_5$) to discriminate between normal and abnormal objects from the UBnormal data set. Results are reported on Avenue and ShanghaiTech. Best results are highlighted in bold.
Legend: * -- results taken from \cite{Georgescu-TPAMI-2021}.
\label{table:results}}
\vspace{-0.2cm}
\end{table} 

\begin{table}[t]
\setlength\tabcolsep{2.2pt}
\small{
\begin{center}
\begin{tabular}{|l|c|c|c|c|}
\hline %
\multirow{2}{*}{Method} &  \multicolumn{2}{|c|}{Avenue}  &  \multicolumn{2}{|c|}{ShanghaiTech}\\ 
\cline{2-5}
&  RBDC      & TBDC   &  RBDC      & TBDC \\
\hline
\hline
Liu \etal~\cite{Liu-CVPR-2018} & $19.59^*$ & $56.01^*$ & $17.03^*$ & $54.23^*$ \\
Ionescu \etal~\cite{Ionescu-CVPR-2019} & $15.77^*$ & $27.01^*$  & $20.65^*$ & $44.54^*$  \\
Ramachandra \etal~\cite{Ramachandra-WACV-2020a} & $35.80$ & $\mathbf{80.90}$ & -  & -\\
Ramachandra \etal~\cite{Ramachandra-WACV-2020b} & $41.20$ & $78.60$ & - & - \\
Georgescu  \etal~\cite{Georgescu-TPAMI-2021} & $\mathbf{65.05}$ & $66.95$ &  $41.34$ &  $78.79$\\  
\hline
\cite{Georgescu-CVPR-2021}  & $57.00$ & $58.30$ &  $42.80$ &  $83.90$\\   
\cite{Georgescu-CVPR-2021} + $T_5$ (ours) & $58.69$ & $59.84$ &  $44.30$ &  $84.56$\\  
\cite{Georgescu-CVPR-2021} + $T_5$ + CycleGAN (ours) & $61.10$ & $61.38$ &  $\mathbf{47.15}$ &  $\mathbf{86.15}$\\  
\hline
\end{tabular}
\end{center}
} 
\vspace{-0.5cm}
\caption{RBDC and TBDC scores (in \%) of the state-of-the-art methods~\cite{Liu-CVPR-2018,Ionescu-CVPR-2019,Ramachandra-WACV-2020a,Ramachandra-WACV-2020b,Georgescu-TPAMI-2021} versus an approach based on the framework proposed in \cite{Georgescu-CVPR-2021}, enhanced by us with the fifth proxy task ($T_5$) to discriminate between normal and abnormal objects from the UBnormal data set. Results are reported on Avenue and ShanghaiTech. Best results are highlighted in bold.
Legend: * -- results taken from \cite{Georgescu-TPAMI-2021}.
\label{table:results_BDC}}
\vspace{-0.2cm}
\end{table}

\begin{figure}[!t]
\begin{center}
\centerline{\includegraphics[width=1.0\linewidth]{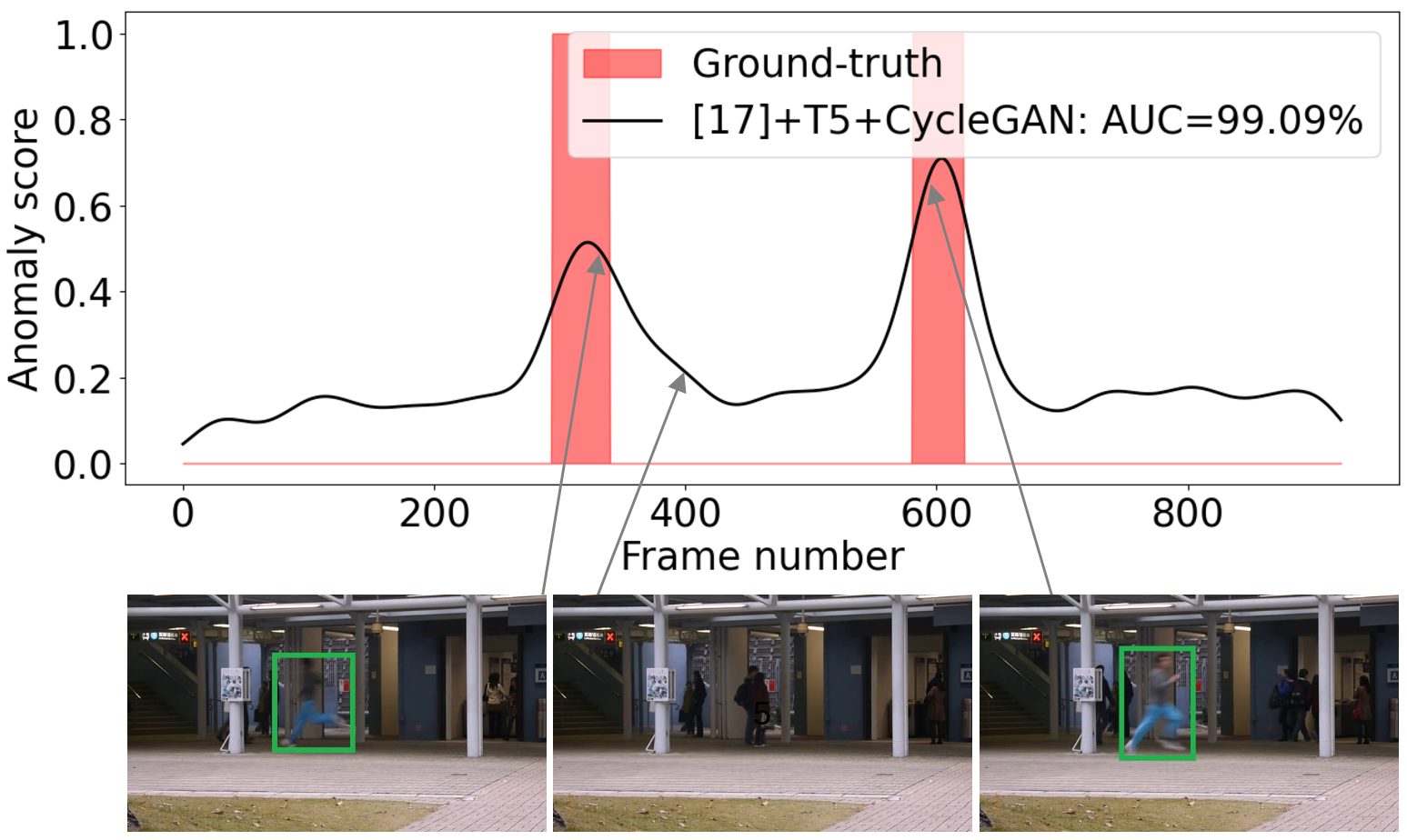}}
\vspace{-0.2cm}
\caption{Frame-level scores and anomaly localization examples for test video $03$ from the Avenue data set. Best viewed in color.}
\label{avenue_03}
\end{center}
\vspace{-0.8cm}
\end{figure}
 
\noindent
\textbf{Avenue.}
We report the results obtained on the CUHK Avenue\cite{Lu-ICCV-2013} data set in Tables~\ref{table:results} and \ref{table:results_BDC}. We compare the approach based on UBnormal samples with the following state-of-the-art methods~\cite{Lee-ICASSP-2018,Liu-CVPR-2018,Liu-BMVC-2018,Sultani-CVPR-2018,Gong-ICCV-2019,Ionescu-CVPR-2019,Ionescu-WACV-2019,Lee-TIP-2019,Nguyen-ICCV-2019,Vu-AAAI-2019,Wu-TNNLS-2019,Dong-Access-2020,Doshi-CVPRW-2020a,Doshi-CVPRW-2020b,Ji-IJCNN-2020,Lu-ECCV-2020,Park-CVPR-2020,Ramachandra-WACV-2020a,Ramachandra-WACV-2020b,Sun-ACMMM-2020,Tang-PRL-2020,Wang-ACMMM-2020,Astrid-BMVC-2021,Astrid-ICCVW-2021,Chang-RP-2022,Georgescu-TPAMI-2021,Madan-ICCVW-2021,Li-CVIU-2021,Liu-ICCV-2021,Szymanowicz-CVPRW-2021,Yang-Access-2021,Yu-TNNLS-2021}. Without closing the distribution gap between UBnormal and CUHK Avenue\cite{Lu-ICCV-2013} via CycleGAN, we still manage to obtain an improvement of $0.4\%$ in terms of the micro-averaged AUC and $0.5\%$ in terms of the macro-averaged AUC compared to the results reported by Georgescu \etal \cite{Georgescu-CVPR-2021} at the object level. Upon passing the objects through CycleGAN, we obtain the state-of-the-art results of $93.0\%$ and $93.2\%$ in terms of micro-averaged and macro-averaged frame-level AUC, surpassing the original method~\cite{Georgescu-CVPR-2021} by at least $1.3\%$ for all four metrics. 

In Figure~\ref{avenue_03}, we present the frame-level anomaly scores and some examples of anomaly localization for test video $03$ from Avenue. We observe that the approach based on five tasks can precisely localize and detect the two anomalies. 

\begin{figure}[!t]
\begin{center}
\centerline{\includegraphics[width=1.0\linewidth]{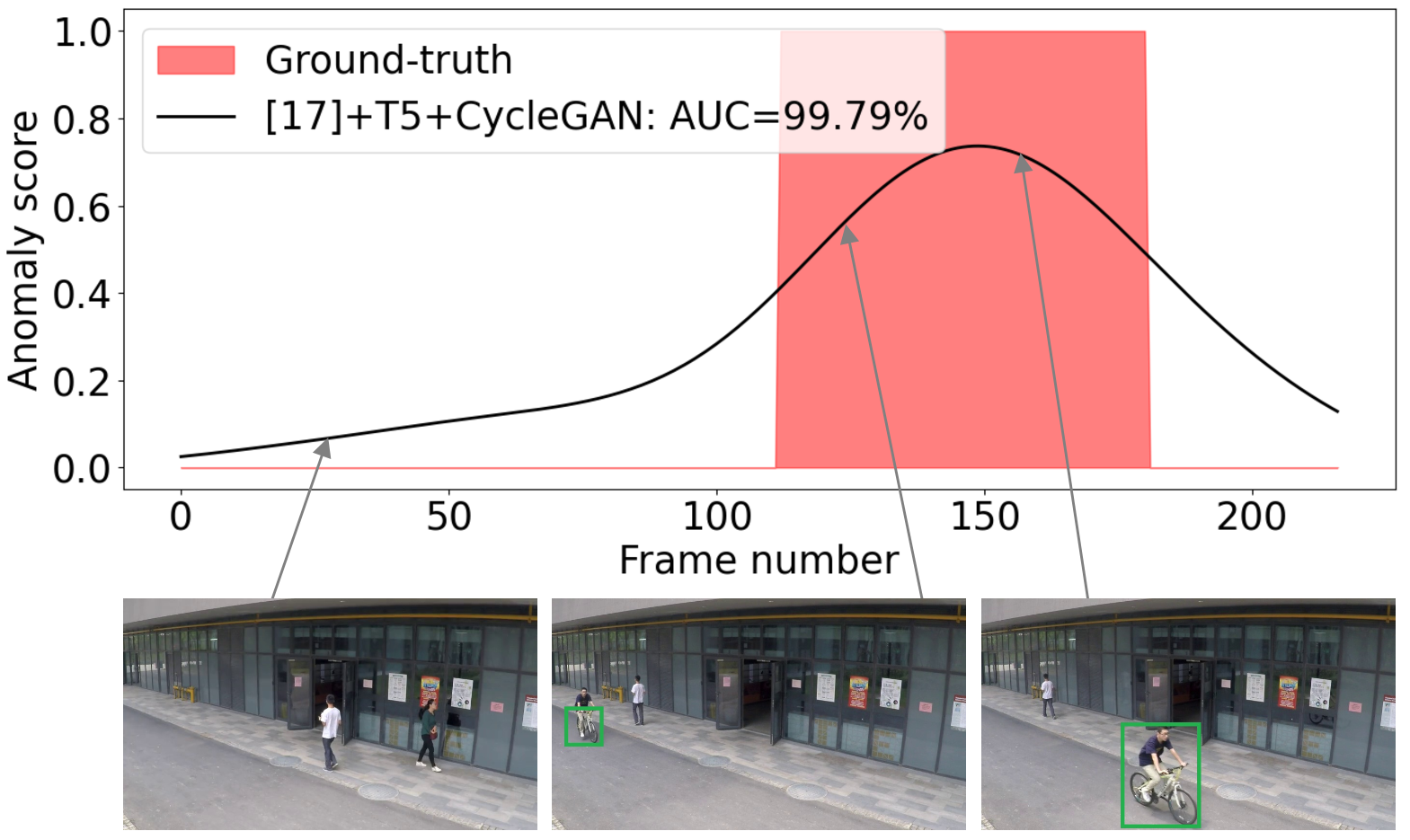}}
\vspace{-0.2cm}
\caption{Frame-level scores and anomaly localization examples for test video $06\_0153$ from ShanghaiTech. Best viewed in color.}
\label{shanghai_06_0153}
\end{center}
\vspace{-0.8cm}
\end{figure}

\noindent
\textbf{ShanghaiTech.} In Tables~\ref{table:results} and \ref{table:results_BDC},
we also report the results obtained on the ShanghaiTech \cite{Luo-ICCV-2017} data set, comparing the approach based on UBnormal samples with other state-of-the-art methods~\cite{Liu-CVPR-2018,Gong-ICCV-2019,Sultani-CVPR-2018,Ionescu-CVPR-2019,Lee-TIP-2019,Dong-Access-2020,Doshi-CVPRW-2020a,Doshi-CVPRW-2020b,Lu-ECCV-2020,Park-CVPR-2020,Ramachandra-WACV-2020a,Ramachandra-WACV-2020b,Sun-ACMMM-2020,Tang-PRL-2020,Wang-ACMMM-2020,Astrid-BMVC-2021,Astrid-ICCVW-2021,Chang-RP-2022,Georgescu-TPAMI-2021,Madan-ICCVW-2021,Li-CVIU-2021,Liu-ICCV-2021,Szymanowicz-CVPRW-2021,Yang-Access-2021,Yu-TNNLS-2021}. The results obtained before applying the CycleGAN model surpass the original object-level method of Georgescu \etal \cite{Georgescu-CVPR-2021} by at least $0.2\%$ for all four metrics. After processing the UBnormal objects with CycleGAN, we obtain the state-of-the-art results of $83.7\%$ and $90.5\%$ in terms of the micro-averaged and macro-averaged frame-level AUC measures. We also surpass the object-level baseline by $4.35\%$ and $2.25\%$ in terms of RBDC and TBDC, respectively. Regardless of how UBnormal data is integrated into the model (with or without domain adaptation), it is the data itself that has the merit of increasing performance.

In Figure~\ref{shanghai_06_0153}, we present the frame-level anomaly scores and an example of anomaly localization for video $06\_0153$ from ShanghaiTech. We observe that the approach based on five tasks can accurately localize and detect the anomaly.

\vspace{-0.1cm}
\section{Conclusion}
\vspace{-0.1cm}

In this work, we introduced UBnormal, a novel benchmark for video anomaly detection. To the best of our knowledge, this is the first and only benchmark for supervised open-set anomaly detection. Perhaps the only limitation of UBnormal is that it is composed of virtual characters and simulated actions. However, we showed several important benefits that justify the importance of UBnormal: $(i)$ it enables the fair and head-to-head comparison of open-set and closed-set models, and $(ii)$ it can alleviate the lack of training anomalies in real-world data sets, bringing significant improvements over the current state-of-the-art models.

The results reported on UBnormal with three state-of-the-art models \cite{Georgescu-TPAMI-2021,Sultani-CVPR-2018, Bertasius-ICML-2021} indicate that our benchmark is very challenging. Nonetheless, we consider that the chosen baselines are strong competitors for future works trying to further improve performance on our benchmark.

In future work, we aim to test the benefit of augmenting other video anomaly detection data sets. Perhaps one of the options that seem more appealing is UCF-Crime \cite{Sultani-CVPR-2018}. However, given that the abnormal classes in UCF-Crime are available at both training and test time, \ie the task is closed, we believe that adding UBnormal data will likely not bring very high performance gains.

\section*{Acknowledgments}

The research leading to these results has received funding from the EEA Grants 2014-2021, under Project contract no.~EEA-RO-NO-2018-0496. This article has also benefited from the support of the Romanian Young Academy, which is funded by Stiftung Mercator and the Alexander von Humboldt Foundation for the period 2020-2022.

{\small
\bibliographystyle{ieee_fullname}
\bibliography{references}
}

\clearpage

\begin{figure*}[!ht]
\begin{center}
\centerline{\includegraphics[width=0.98\textwidth]{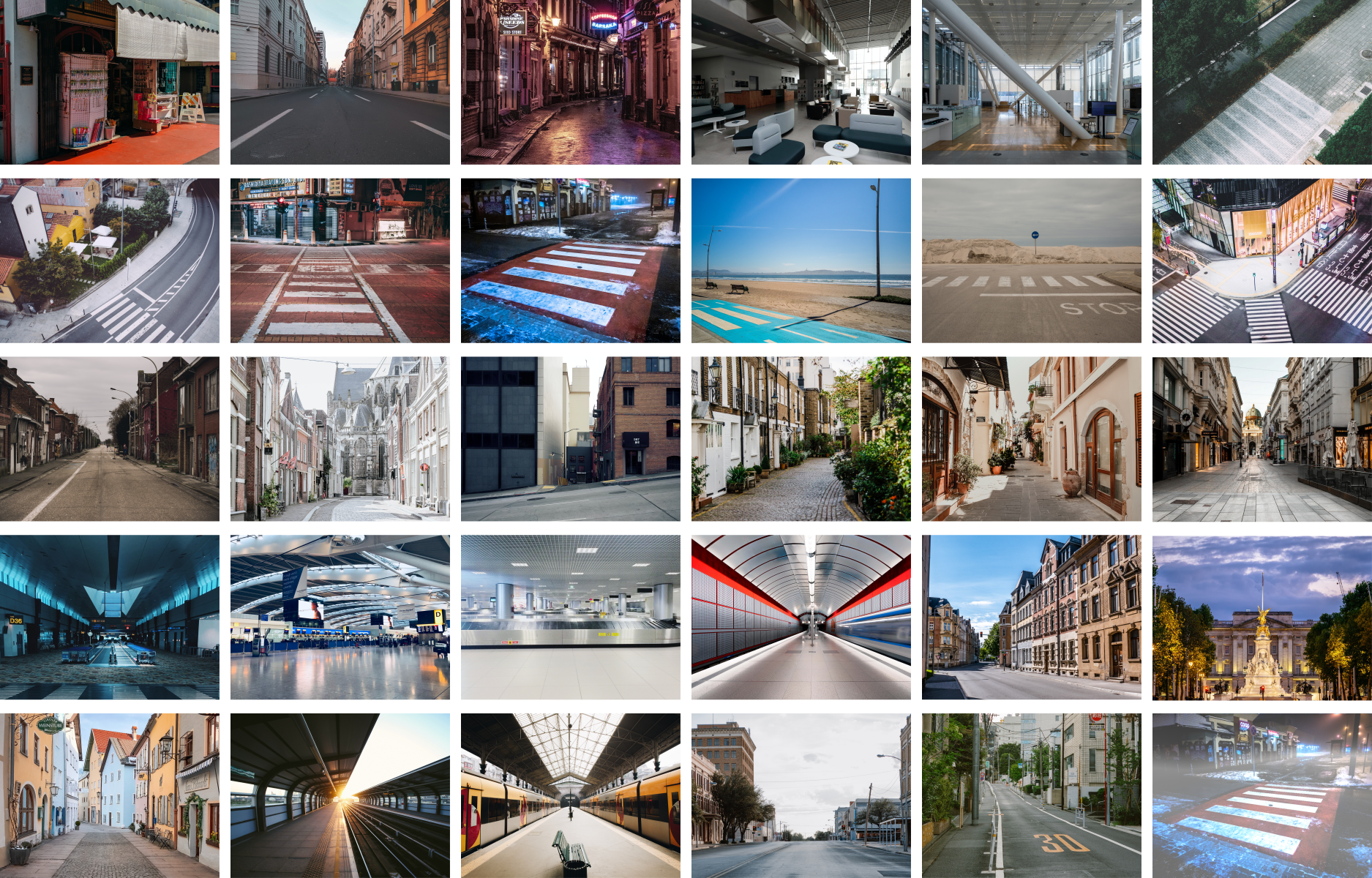}}
\caption{The 29 backgrounds used to generate the virtual scenes in UBnormal. One of the scenes is repeated (in the bottom-right corner) to show the fog effect.  Best viewed in color.}
    \label{fig:scenes}
\vspace{-0.5cm}
\end{center} 
\end{figure*}

\begin{figure*}[!ht]
\begin{center}
\centerline{\includegraphics[width=0.95\linewidth]{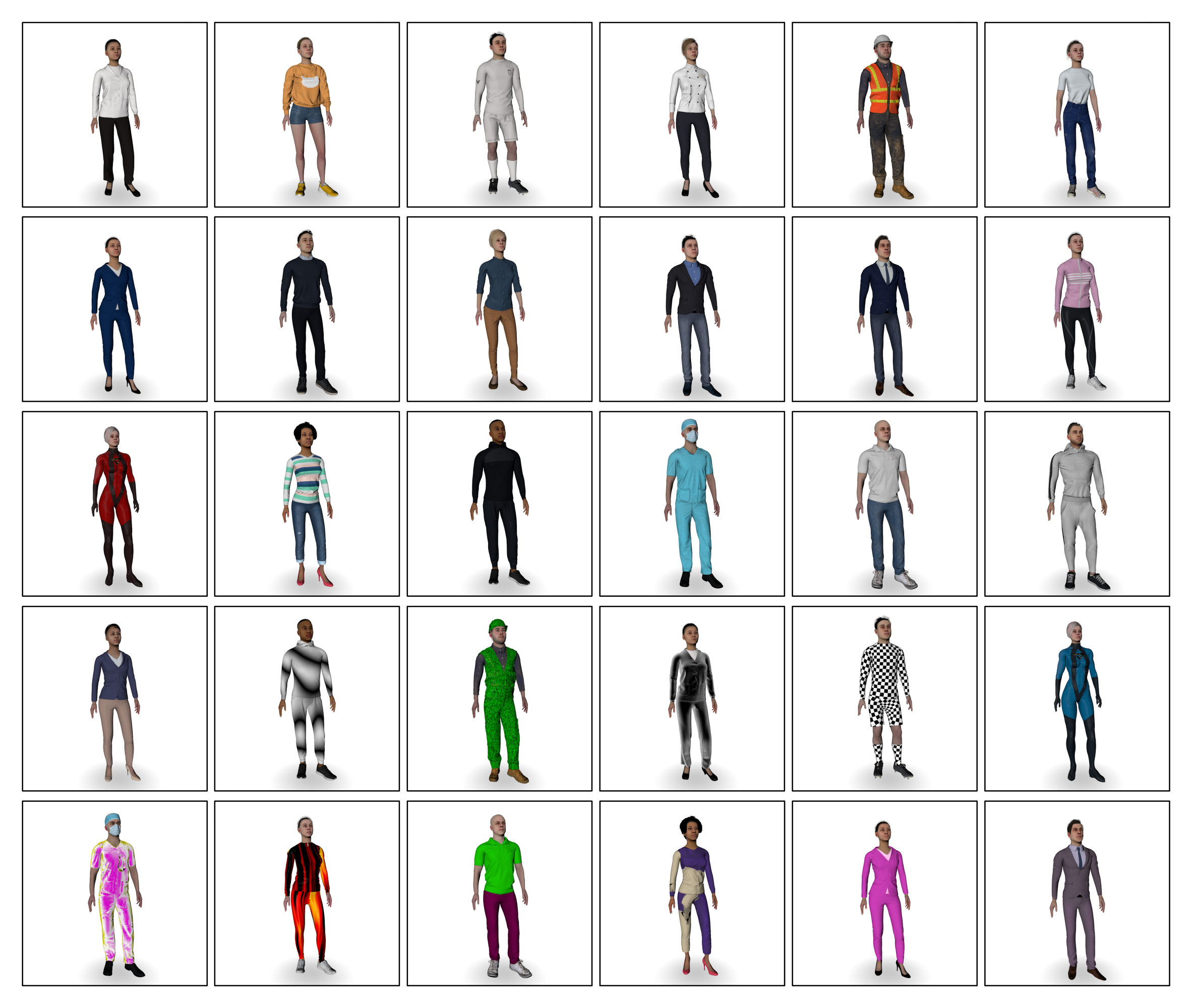}}
\vspace{-0.4cm}
\caption{Various characters animating the scenes in the UBnormal benchmark. There are $19$ unique characters that serve as seeds for generating people with different hair color and clothes. Best viewed in color.}
\label{fig:characters}
\vspace{-0.5cm}
\end{center} 
\end{figure*}

\section{Supplementary}

\begin{figure*}[!ht]
\begin{center}
\centerline{\includegraphics[width=0.82\linewidth]{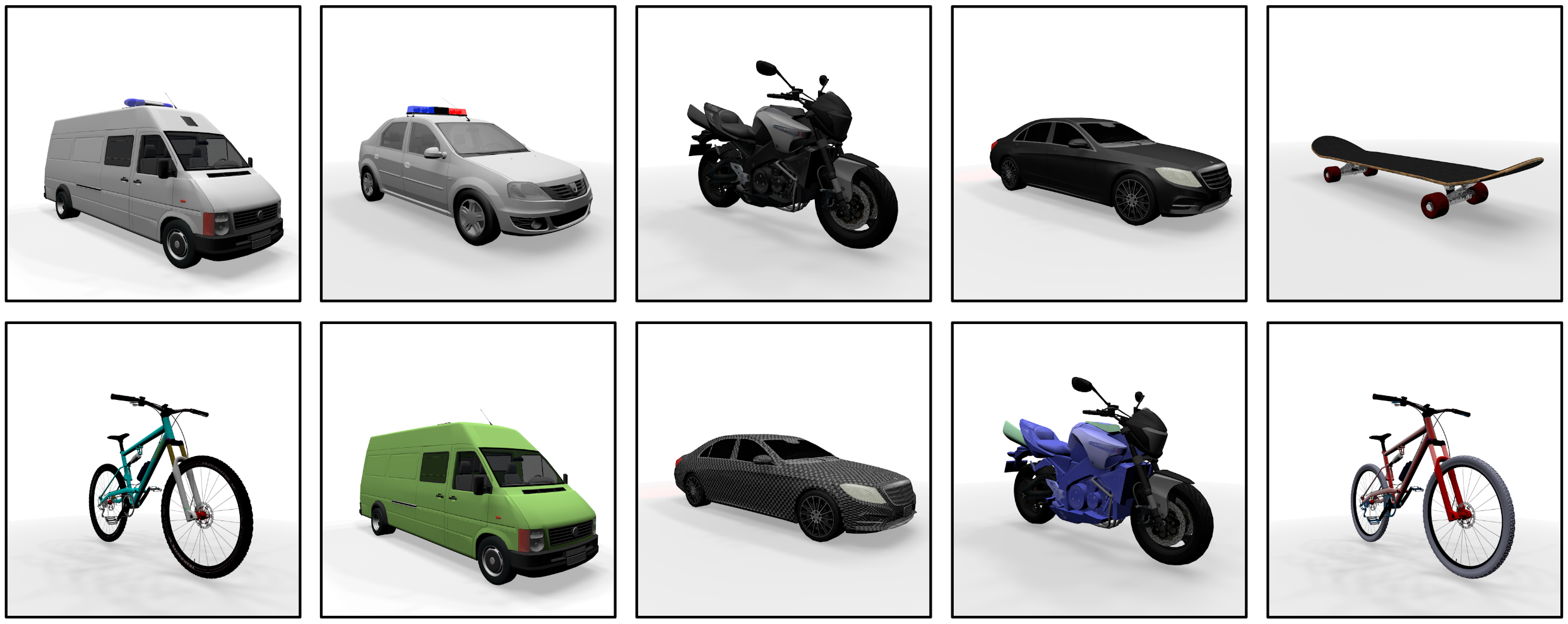}}
\vspace{-0.2cm}
\caption{Various objects animating the scenes in the UBnormal benchmark. There are $5$ object categories besides people. To increase variation, we apply different colors to the objects. Best viewed in color.}
\label{fig:objects}
\end{center} 
\end{figure*}

\begin{figure*}[!ht]
\begin{center}
\centerline{\includegraphics[width=0.92\linewidth]{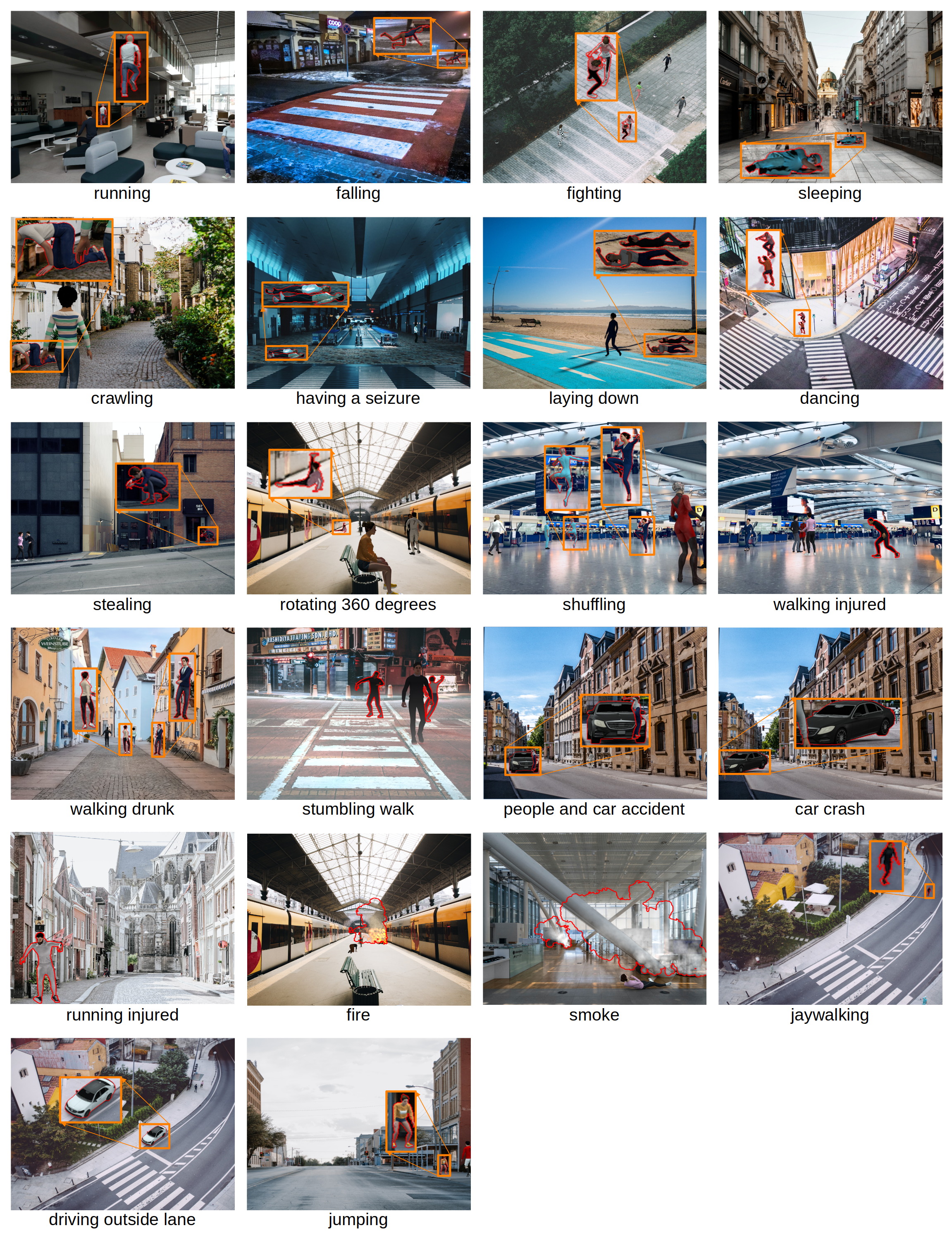}}
\vspace{-0.3cm}
\caption{The abnormal action categories from UBnormal. The abnormal objects are emphasized through a red contour. To improve readability, we apply a magnifying effect to smaller objects. Best viewed in color.}
\label{fig:anomalies}
\end{center} 
\end{figure*}

\subsection{Visualizations}

In Figure~\ref{fig:scenes}, we present the $29$ scenes of our data set. We have $22$ outdoor scenes and $7$ indoor scenes. There are various illumination conditions, corresponding to different weather conditions (\eg sunny, cloudy, foggy) and different day times (\eg day, sunset, night). The last image (at the bottom right) illustrates one of the existing scenes with the fog effect. 

We present the animated characters from the UBnormal data set in Figure~\ref{fig:characters}. There are $19$ unique characters, but in order to increase the variety of the data set, we augment the characters by changing the color of their hair and clothes.  

In Figure~\ref{fig:objects}, we show the $5$ object categories (excluding people) from UBnormal. We variate the colors of objects to increase diversity, as for the animated characters.

In Figure~\ref{fig:anomalies}, we illustrate an example for each abnormal action category from the UBnormal benchmark. There are $20$ anomaly types related to objects (\eg people, cars) and two non-object anomaly types (fire, smoke). The figure shows the diversity of our anomalies.

\begin{table}[t]
\setlength\tabcolsep{3.9pt}
\small{
\begin{center}
\begin{tabular}{|l|r|r|r|r|}
 \hline 
 {Statistics}  &  \multicolumn{1}{|c|}{{Training}}   &   \multicolumn{1}{|c|}{{Validation}}   & \multicolumn{1}{|c|}{{Test}} &  \multicolumn{1}{|c|}{{Total}}\\
\hline 
\hline
$\#$anomalies   &       $195$      &	$76$	         & $389$    &  $660$ \\
\hline
$\#$abnormal frames &    $25,\!227$   &	$13,\!938$	     & $49,\!850$  &  $89,\!015$ \\
\hline
$\#$normal frames   &   $90,\!860$    &	$14,\!237$	     & $42,\!790$  &  $147,\!887$ \\
\hline  
$\#$abnormal minutes &   $14.02$   &	$7.74$	         & $27.69$  &  $49.45$ \\
\hline
$\#$normal minutes  &   $50.48$    &	$7.91$	         & $23.77$  &  $82.16$ \\
\hline
$\#$videos            &  $268$        &	$64$	         & $211$   &  $543$ \\
\hline
$\#$abnormal regions &  $38,\!048$    &	$21,\!321$	     & $82,\!738$  &  $142,\!107$ \\
\hline
$\#$unique objects   &  $1,\!443$     &	$351$	         & $1,\!114$   &  $2,\!908$ \\
\hline
avg. objects / frame    &  $4.58$     &	$4.53$	         & $4.24$   &  $4.44$ \\
\hline
\end{tabular}
\end{center}
} 
\vspace{-0.55cm}
\caption{Detailed statistics for the UBnormal data set. Our videos are generated at $30$ FPS.\label{table:statistic_our_db}}  
\vspace{-0.45cm}
\end{table}

\subsection{Statistics}

In Table~\ref{table:statistic_our_db}, we report several statistics about the UBnormal data set. The number of videos in the data set is $543$, with $268$ videos for training, $64$ for validation and the remaining $211$ for testing. There are a total of $660$ anomalies divided into $195$, $76$ and $389$ for training, validation and test, respectively. The total number of abnormal frames is $89,\!015$, with the largest fraction of $49,\!850$ frames belonging to the test set. There are $142,\!107$ abnormal regions in the data set, which accounts for an average of $1.60$ abnormal regions per abnormal frame. The average number of objects per frame is $4.44$. The overall length of our data set is $132$ minutes.

\end{document}